\UseRawInputEncoding
\documentclass[journal=jacsat,manuscript=article]{achemso}
\usepackage[version=3]{mhchem} 
\usepackage{listings}
\usepackage{xcolor}
\usepackage{hyperref}

\lstdefinestyle{mypython}{
    language=Python,
    basicstyle=\ttfamily\small,
    keywordstyle=\color{blue}\bfseries,
    stringstyle=\color{red},
    commentstyle=\color{gray}\itshape,
    numbers=left,
    numberstyle=\tiny,
    stepnumber=1,
    numbersep=5pt,
    breaklines=true,
    frame=single,
    showstringspaces=false,
    captionpos=b
}

\author{Peter Pak}
\affiliation{
  Department of Mechanical Engineering, Carnegie Mellon University, Pittsburgh,
  PA, USA
}
\author{Achuth Chandrasekhar}
\affiliation{
  Department of Mechanical Engineering, Carnegie Mellon University, Pittsburgh,
  PA, USA
}
\author{Amir Barati Farimani}
\email{barati@cmu.edu}
\affiliation{
  Department of Mechanical Engineering, Carnegie Mellon University, Pittsburgh,
  PA, USA
}
\alsoaffiliation{
  Machine Learning Department, Carnegie Mellon University, Pittsburgh, PA, USA
}

\title[]{Agentic Additive Manufacturing Alloy Evaluation}

\abbreviations{IR,NMR,UV}

\SectionNumbersOn
\begin{document}

\begin{abstract}
Agentic systems enable the intelligent use of research tooling, augmenting a
researcher's ability to investigate and propose novel solutions to existing
problems. Within Additive Manufacturing (AM), alloy selection and evaluation remains a complex
challenge, often requiring expertise in the various domains of materials
science, thermodynamic simulations, and experimental analysis. Large Language
Model (LLM) enabled agents can facilitate this endeavor by utilizing their
extensive knowledge base to dispatch tool calls via Model Context Protocol (MCP)
to perform actions such as thermophysical property diagram calculations and lack of
fusion process map generation. In addition, the multi-agent system can effectively reason through complex user prompts and provide
analysis on the lack of fusion process window of common alloys such as SS316L and IN718 along with proposed composition variants of known alloys. These agents can dynamically
adjust their task trajectory to the outcomes of tool call results, effectively
enabling autonomous decision-making in practical environments. This work aims to showcase the benefits
of adopting a LLM enabled multi-agent system to automate and accelerate the task of evaluating proposed additive manufacturing alloys, both novel and known.
\end{abstract}

\section{Introduction}
Evaluation of an alloy's suitability for fabrication and its appropriate processing parameters is a key component of the Additive Manufacturing (AM) process.
In industries that focus on biomedical, aerospace, or energy challenges, their unique applications often involve the selection and evaluation of alloy candidates that are best equipped to perform in their service environment\cite{bandyopadhyay_additive_2018, bandyopadhyay_alloy_2022,
ghafarollahi_atomagents_2024, tammas-williams_design_2017,
polonsky_closing_2020, sames_metallurgy_2016}. In addition, considerations such
as deformation\cite{jadhav_stressd_2023, rashid_review_2023, paul_effect_2014},
corrosion resistance\cite{bandyopadhyay_additive_2018, xu_corrosion_2020}, and
biocompatibility\cite{pesode_additive_2022, pesode_additive_2023,
jiao_additive_2022} are key motivators for composition refinement which can foster the development of novel
alloys. However, the discovery and validation of new AM alloys remains a
time-consuming process that often requires expertise in materials science,
computational simulations, and experimental
analysis\cite{bandyopadhyay_alloy_2022, tammas-williams_design_2017,
polonsky_closing_2020}. Furthermore, each alloy presents its own set of unique
challenges, often requiring specific build parameters to avoid potential defects
within part fabrication\cite{bandyopadhyay_alloy_2022, ghanadi_effect_2025,
edwards_electron_2013, vrancken_heat_2012, jelis_metallurgical_2015,
popovich_microstructure_2016, lu_microstructure_2015, huang_microstructure_2025,
martendal_mitigating_2025}. The search and optimization of desirable process
parameters often requires extensive simulation analysis and  experimental
trials to validate their suitability within build
conditions\cite{rashid_review_2023, honarmandi_rigorous_2021,
tang_prediction_2017, tang_rapid_2016, tan_graded_2015}.

For the task of obtaining material properties for a novel elemental composition, Computer Calculation of Phase
Diagrams\cite{kaufman_computer_1970} (CALPHAD) is commonly performed to
calculate the individual phases of an alloy via Gibbs Free Energy and numerical
optimization. CALPHAD provides a rigorous thermodynamic framework for predicting
equilibrium phase stability and composition in multicomponent alloys by
combining assessed Gibbs energy descriptions of individual phases with numerical
minimization of the total free energy. In practice, modern CALPHAD workflows go
beyond phase fields, as well-curated mobility and property databases enable the
prediction of temperature-dependent thermophysical quantities such as thermal
conductivity ($k$), specific heat capacity ($C_p$), and density
($\rho$).\cite{gheribi_application_2012, campbell_development_2014}.
An estimation of material properties for a proposed alloy composition is predicted with this method using a
library of resources to obtain relevant information regarding various elements,
alloys, and application properties.

With the procurement of the material properties of a proposed alloy composition,
its suitability for an AM build is assessed through numerical solvers. Solvers
such as those developed by Eagar-Tsai\cite{eagar_temperature_1983} and
Rosenthal\cite{rosenthal_theory_2022} provide preliminary information on the
temperature field of a melt pool through an analytical solution. Whereas,
OpenFOAM\cite{noauthor_openfoam_2025} and FLOW-3D\cite{noauthor_flow-3d_nodate}
utilize Computational Fluid Dynamics (CFD) to provide a more detailed analysis
of the underlying fluid flow, heat transfer, and solidification phenomena of the
melt pool\cite{zhang_modeling_2019, strayer_accelerating_2022}. The obtained
melt pool dimensions are used to calculate potential defect regimes within a
process map of beam power and scanning velocity
combinations\cite{cunningham_analyzing_2017, cunningham_defect_2018,
tang_prediction_2017}. Specifically, the lack of fusion defect region is of
primary concern as insufficient melting induces the formation of large pores and
in extreme cases can cause the build process to fail\cite{tang_prediction_2017}.
Insight into this defect regime is essential for the informed selection of
optimal build parameters within the process window.

Recent advances in multi-agent systems have demonstrated the potential of Large
Language Models (LLMs) to perform complex, goal-oriented tasks that extend
beyond the constraints of a single prompt\cite{jadhav_llm-3d_2025,
ock_catalyst_2023, ock_large_2025, chaudhari_modular_2025, george_llm_2025,
bartsch_llm-craft_2025, chandrasekhar_automating_2025, merrill_llm-drone_2025}.
This has been applied to the monitoring and adjustment of real-time Fuse
Deposition Modeling (FDM) builds\cite{jadhav_llm-3d_2025}, search and discovery
of potential catalyst and drug candidates\cite{ock_catalyst_2023,
ock_large_2025}, and optimization of material and molecular design
simulations\cite{chaudhari_modular_2025, chandrasekhar_automating_2025}. In
these tasks, LLMs autonomously determine the next course of action by reasoning
over the information available within the dynamic
environment\cite{jadhav_llm-3d_2025, chaudhari_modular_2025, ock_large_2025}.
The next course of action often involves utilizing functionality external to the
LLM which is made accessible through a tool call. This can be done through
various methods and solution providers include
LangChain\cite{noauthor_langchain_nodate} and
LlamaIndex\cite{noauthor_llamaindex_nodate}. This work utilizes the Model
Context Protocol (MCP)\cite{anthropic_introducing_2024} as the library for
creating and exposing tools for the LLM that allow seamless integration into
any client with an MCP interface.

With these established goals, this work aims to develop a multi-agent system for
the search, proposal, and analysis of novel alloy compositions suitable for
additive manufacturing. The inclusion of the LLM allows for the intelligent automation and reasoning of tool generated responses from natural language
inputs which can be constructed into valid function inputs. MCP
tool binding from \texttt{thermo-calc} and \texttt{additive-manufacturing}
packages allow for the actions such as process map generation and material
property prediction to be controlled by agents. The multi-agent system is able
to reason and plan through given tasks and dynamically adjust task trajectories
based on tool responses. This in turn enables the intelligent automation of routine
tasks and accelerates the evaluation of novel alloy compositions within additive
manufacturing.

\begin{figure}[htbp]
    \centering
    \includegraphics[width=\textwidth]{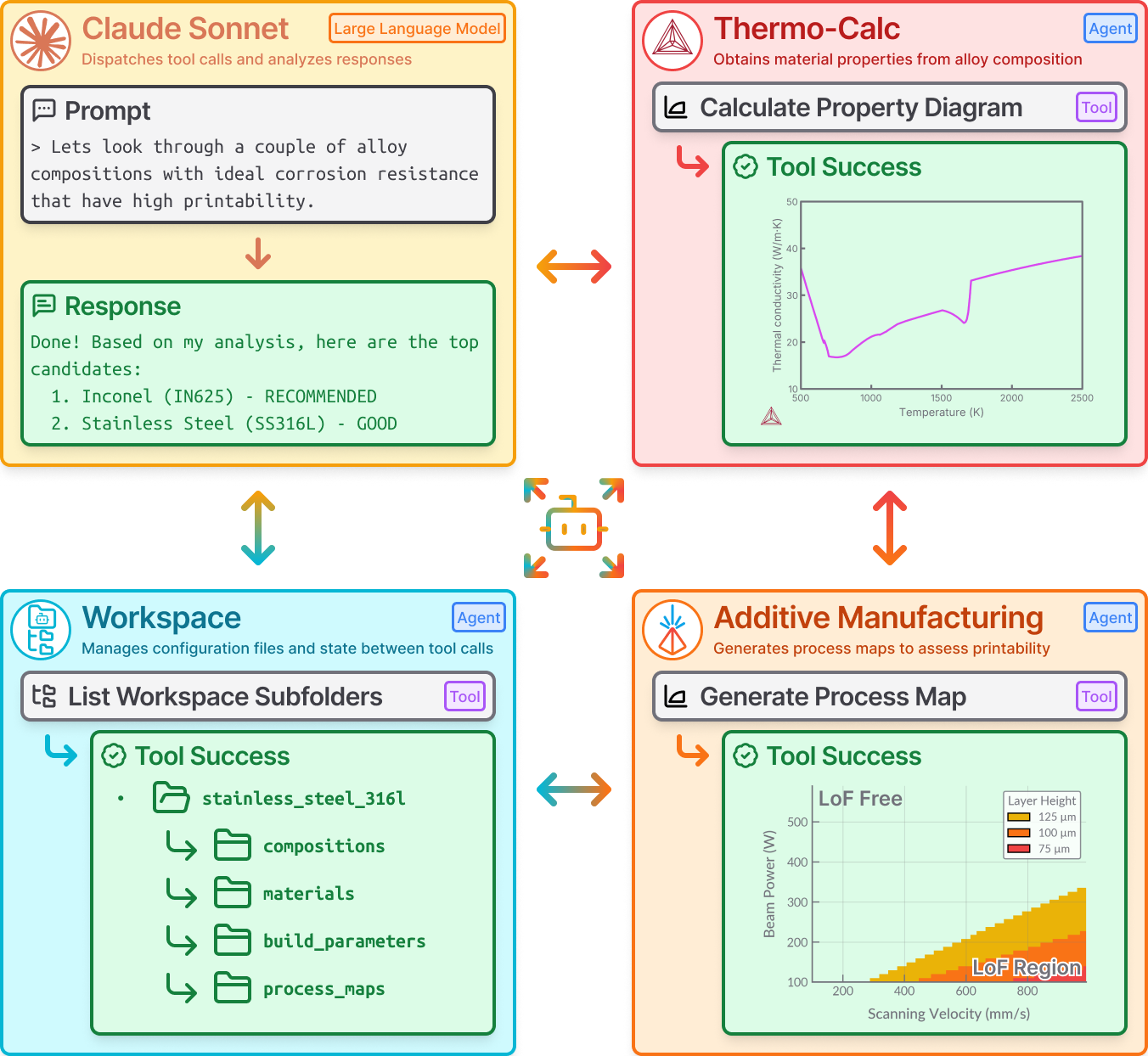}
    \caption{
    \textbf{(Top Left)} An input query for alloy compositions regarding the
    printability of an additively manufactured part suitable for its intended
    use case is provided to Claude Sonnet. This Large Language Model (LLM) calls
    the tools necessary to generate and analyze each potential alloy
    compositions, providing a response of candidates ranked by their content of
    their lack of fusion fusion regimes. \textbf{(Top Right)} Thermo-Calc allows
    for the retrieval of material properties for an hypothetical alloy composition,
    for instance thermal conductivity, to be used in down stream lack of fusion
    calculations. \textbf{(Bottom Left)} Workspaces provide a way for each of
    the tools to effectively communicate with one another and handles state
    management and file organization. \textbf{(Bottom Right)} Tools managed by
    the Additive Manufacturing subagents then utilize the calculated material
    properties from the Thermo-Calc subagent to generate a lack of fusion
    process map to send back to the LLM for analysis and final recommendation.
    }
    \label{fig:main}
\end{figure}

\section{Methodology}

\subsection{Calculation of Thermophysical Properties}
\label{sec:thermo_calc_calcuation}
For this task, given an hypothetical element composition of a specific alloy,
Thermo-Calc\cite{thermo-calc_software_thermo-calc_2025} is expected to provide
the relevant material properties of density, thermal conductivity, specific heat
capacity, electric resistivity, and the liquidus and solidus phase transition
temperatures. Thermo-Calc utilizes a CALPHAD based solver to calculate
equilibrium phase diagrams and extrapolate properties along with the material
property databases to assist with obtaining thermophysical properties for
various alloys. The platform also includes their TC-Python SDK allowing for
programmatic use of the features, providing an interface for the creation and
binding of agentic tools callable via MCP. 

With the Thermo-Calc's property diagram calculation function, a composition of
elements is provided to generate a property diagram to extract material
properties. Compositions consist of the mass fraction of individual elements and
for commonplace alloys, a map of their elemental compositions is obtained from
existing literature sources \cite{akbari_machine_2024, lober_selective_2014,
manfredi_powders_2013, jelis_metallurgical_2015, sun_evaluation_2015,
yadollahi_effects_2015, popovich_microstructure_2016, kok_anisotropy_2018,
schmidtke_process_2011, hack_mechanical_2017, rodriguez_dynamic_2015,
lu_microstructure_2015, tan_graded_2015, vrancken_heat_2012,
edwards_electron_2013}. In this process, a suitable database (one of
\texttt{TCFE14}, \texttt{TCNI12}, \texttt{TCAL9}, \texttt{TCTI6}, and
\texttt{TCHEA7}, or \texttt{PURE5} \cite{calphad_textbook}) is selected to
obtain the necessary phase and transport properties with the appropriate SI
units. Database selection is determined through matching an alloy's primary
element composition to its database counterpart. The element with the highest weight fraction is designated as the "top" element. A composition is deemed as multi-principal if at least three elements have individual weight fractions \(\ge 0.15\). Then the appropriate database is selected by elimination in the order as described in Appendix C.
The provided compositions and database values are then utilized in equilibrium calculations
where the minimization of Gibbs Free Energy is iteratively calculated until
convergence is met \cite{noauthor_cross_2023, noauthor_property_nodate}.

Solidus and liquidus phase transition temperatures are extrapolated through
tracking the liquid volume fraction. A one-dimensional temperature sweep is
performed to find a point where the liquid fraction moves from approximately 0
to 1. The default arguments for the minimum and maximum temperature ranges are
500 K and 3500 K respectively. The obtained solidus and liquidus temperatures
are utilized in subsequent tasks to obtain material properties from the proposed
alloy. Density ($\rho = \mathrm{B}/\mathrm{V}$) and specific heat capacity are
obtained through Thermo-Calc's user function using the ratio of mass
(\texttt{B}) to volume (\texttt{V}) and the temperature derivative of molar
enthalpy denoted as \texttt{HM.T} respectively. Thermal conductivity and
electrical resistivity are built-in thermodynamic quantities and can be readily
obtained at a given temperature. The melting temperature is taken as the average
between the solid and liquidus phases.

The calculated material properties of absorptivity, thermal conductivity,
liquidus and solidus phase transition temperatures, density, and specific heat
capacity are saved as a common material configuration compatible with the
\texttt{additive-manufacturing} package. These values will then be utilized with
the tools there for initialization and generation of a lack of fusion process
map to evaluate the printing feasibility of these material properties.

The accuracy of the material properties thus obtained, are highly dependent on the coverage and fidelty of the Thermo-Calc databases used. For complex alloy chemistries, such as multi-principal element alloys the software may have to interpolate between sparse input data points or extrapolate beyond them. Moreover, the present study does not utilize independently calibrated models and the quantitative accuracy is determined by good agreement with literature trends. All of these factors may introduce some uncertainty in the final predictions.

\subsubsection{Absorptivity Calculation and Limitation}
Absorptivity is approximated using the series expansion for emissivity in a
direction normal to the surface, based on Drude's Theory (Equation \ref{eq:bramson_original}).
\cite{bramson_infrared_1968} Here, $\varepsilon_\lambda\left(T\right)$ denotes
the emissivity at a given wavelength ($\lambda$) and temperature ($T$) where the
conductivity of a metal ($\gamma$) is expected in units of $(\Omega \cdot
\mathrm{cm})^{-1}$.

\begin{equation}
\varepsilon_\lambda\left(T\right) = \frac{0.365}{\sqrt{\gamma \lambda}} -
\frac{0.0667}{\gamma \lambda} +
\frac{0.006}{\sqrt{\left(\gamma\lambda\right)^{3}}} - \;...
\label{eq:bramson_original}
\end{equation}

\noindent Since resistivity ($\rho$) and conductivity ($\gamma$) are reciprocal
properties ($\gamma = \frac{1}{\rho}$), Equation \ref{eq:bramson_original} can
be modified to accept our computed electrical resistivity value and near
infrared wavelength of 1070 nm. (Equation \ref{eq:bramson_modified}).

\begin{equation}
\varepsilon_\lambda\left(T\right) = 0.365\sqrt{\frac{\rho}{\lambda}} -
0.0667\frac{\rho}{\lambda} + 0.006\sqrt{\left(\frac{\rho}{\lambda}\right)^{3}} -
\;...
\label{eq:bramson_modified}
\end{equation}

This approximation provides a simple estimation of a given alloy's absorptivity
and serves a baseline value to use with melt pool calculations. A more
comprehensive approach would include the absorptivity value's dependence to
applied power and the reflections observed within a melt pool during
keyholing\cite{trapp_situ_2017, nordet_absorptivity_2022}. For alloys such as
copper which exhibits a large range of absorptivity values (around 5\% to
90\%)\cite{nordet_absorptivity_2022} depending on power and laser conditions,
the predicted results may be limited by the stated assumptions. Our approach is validated by comparison with experimental data in Appendix D\cite{cook_determining_2023, honda_measurement_2025, mills_recommended_2002, he_powder_2022}.

\subsection{Lack of Fusion Defect Prediction}
\label{sec:defect_prediction}
Defects created within the laser powder bed fusion process can arise through
various means and affect the final part's material and mechanical
properties\cite{gordon_defect_2020, cunningham_analyzing_2017,
ng_porosity_2009}. This includes defects such as porosity, microstructual
inhomogenity, and inclusions which can result in degraded performance in fatigue
life as well as mechanical strength\cite{gordon_defect_2020,
bellini_additive_2021}. The source of these defects can be attributed to at
least one of the following process map defect regimes of either Lack of Fusion
(LoF), Keyholing, or Balling\cite{gordon_defect_2020, cunningham_defect_2018}.
Of these defects, porosity lack of fusion porosity is often larger than that generated by keyholing or balling \cite{shrestha_formation_2022}. Pores presents a significant risk to
the fatigue life as it provides a starting point for cracks to nucleate
from\cite{bellini_additive_2021, smith_relationship_2019}.

The criterion for lack of fusion is primarily concerned with overlap between
subsequent melt pool tracks, where adjustments to the hatch spacing along with
layer height determine the extent of unfused powder within the scan track.\cite{gordon_defect_2020, tang_prediction_2017} This is modeled by with
Equation \ref{eq:lof}\cite{tang_prediction_2017, gordon_defect_2020,
pak_additivellm_2025} where computed ratios greater than 1 are expected to
exhibit lack of fusion defects.

\begin{equation}
\left(\frac{Hatch\;Spacing}{Melt\;Pool\;Width}\right)^2 +
\left(\frac{Layer\;Height}{Melt\;Pool\;Depth}\right)^2 \le 1
\label{eq:lof}
\end{equation}

Hatch spacing and layer height are independent process parameters that are
prescribed for the build process. Melt pool dimensions of depth and width are
physical values that obtained from either cross-sectional measurements of the
scan track\cite{aboulkhair_formation_2016} or modeling using solvers such as
that of the Rosenthal\cite{tang_rapid_2016, tang_prediction_2017,
rosenthal_theory_2022} equation (Equation \ref{eq:rosenthal}). 

\subsubsection{Dimensional Approximation of the Melt Pool}

In Equation \ref{eq:rosenthal}, the local temperature $T$ (K) is obtained at a
distance $z$ (m) along the travel direction for a radial distance $R$ (m) from
the beam position\cite{tang_rapid_2016, rosenthal_theory_2022}. Additional
factors such as the temperature of the plate $T_\infty$ (K), applied power $Q$
(W), scan speed $V$ (m/s), thermal diffusivity $\alpha$ ($\text{m}^2$/s), and
thermal conductivity $k$ (W/mK) utilized to provide an approximate calculation
of the temperature field\cite{tang_rapid_2016, rosenthal_theory_2022}. Thermal
diffusivity ($\alpha = \frac{k}{\rho C_p}$) can be obtained from thermal
conductivity $k$ (W/mK), density $\rho$ (kg/$\text{m}^3$), and specific heat
capacity $C_p$ (J/kg$\cdot$K). Radial distance $R$ is the combined coordinate of
$R^2 = z^2 + r^2$ where $z$ (also $-\xi$) is the distance along the center line.

\begin{equation}
z = R + \frac{2 \alpha}{V}ln\left(\frac{2 \pi k R \Delta T}{\epsilon P}\right)
\label{eq:rosenthal_bounds}
\end{equation}

Given the liquidus temperature of a material, Equation \ref{eq:rosenthal} can be
rearranged to provide the bounds of the melt pool. This is shown with Equation
\ref{eq:rosenthal_bounds}, where $\Delta T = T_{melting} - T_{initial}$ and
$\epsilon$ represents the dimensionless absorptivity value. The length of the
melt pool can be obtained by calculating the length of the temperature field in
front and tailing the heat source. The tailing length (Equation
\ref{eq:rosenthal_length_8}) of the heat source can be obtained from setting the
$z$ to $R$ and then solving for $R$ as seen with its derivation included in
\ref{apx:melt_pool_length}. This provides a stop point when passing $R$ values
into Equation \ref{eq:rosenthal_bounds} with a step size of 1 $um$ when
calculating the bounds, recording the maximum to use as the melt pool
dimensions.

\begin{equation}
T = T_\infty + \frac{Q}{2 \pi k R} \exp\left( \frac{V(z - R)}{2\alpha} \right)
\label{eq:rosenthal}
\end{equation}

\subsubsection{Model Assumptions and Limitations}

With Rosenthal's approximation of a moving heat source
a number of assumptions are made which include a melt pool in conduction
mode, a point heat source, and temperature-independent thermal
properties\cite{tang_prediction_2017, rosenthal_theory_2022}. The derived melt
pool dimensions are limited to a steady state conduction mode melt pool and when
applied to the lack of fusion criterion it does not consider dimensional
fluctuations of the melt pool that are present in experimental
builds\cite{li_laser_2023}. 

Since these melt pool dynamics are obtained through a purely conduction based
approach, factors such as keyholing are not reflected in the obtained dimensions
throughout the process map. When viewed through the lens of defects pertaining
to lack of fusion, these dimensions present more conservative estimates to the
process window. This is from the smaller depth estimations of the melt pool in
conduction mode as to that of a keyholing melt pool which has an elongated lower
half\cite{tapia_gaussian_2018, gordon_defect_2020, bellini_additive_2021}. Lack
of fusion and keyhole defect regimes do have some degree of overlap however
experimental data shows this occurs in combinations of low to medium power and
low velocity ranges (Figure \ref{fig:keyhole_lof_boundary}). Energy density can
be used as an indicator for the transition between keyhole and conduction mode
and for alloys with known energy density transition points, this value is used
as a cutoff for conduction based melt pool dimensional approximations
\cite{bellini_additive_2021}.

\subsubsection{Defect Domain Scope and Applicability}
Although this work primarily focuses upon the prediction of lack of fusion
defects within a range of power and velocity combinations, it is worth
addressing the other two common defect regimes of keyholing and balling.

Keyholing describes the elongated depth of the melt pool as it exits conduction
mode, leading to potential defects that can also contribute to lower mechanical
strength in the resulting build\cite{bellini_additive_2021}. A common criterion
for keyholing analyzes the width to depth ratio of the melt pool such that a
melt pool with a depth 1.5 larger than the width is considered to be keyholing
(Equation \ref{eq:keyhole}) \cite{zhang_efficient_2021}.

\begin{equation}
\frac{Width}{Depth} > 1.5
\label{eq:keyhole}
\end{equation}

However, in order to properly model keyhole behavior within the melt pool, a
computational fluid dynamics approach is required to capture effects such as
solid/liquid interfaces and the driving force of recoil pressure from material
vaporization\cite{ki_modeling_2002}. Solvers from
FLOW-3D\cite{noauthor_flow-3d_nodate} and OpenFOAM\cite{noauthor_openfoam_2025}
provide the capability to model these fluid dynamics but present a non-trivial
challenge to integrate into this agentic system as they require both significant
time and computational cost.

Balling presents another potential source of defects which often occurs at high
power and velocity combinations, the grooves generated from the capillary forces
initiate voids within the build if not remelted in a subsequent
pass\cite{gu_balling_2007}. A simplification of the criterion for balling is
comparing the length to width ratio of the melt pool to a set threshold, such as
$\pi$, such that any ratio greater would indicate
balling\cite{zhang_efficient_2021} (Equation \ref{eq:balling}).

\begin{equation}
\frac{Length}{Width} < \pi
\label{eq:balling}
\end{equation}

For the case of balling, a threshold based on the ratio length and width of the
melt pool is rather simplistic and presents an assumption that this behavior
occurs based on the dimensions of the melt pool. Whereas, factors such as the
temperature balance between the melt pool and the solid surface provides a more
grounded approach to capturing the underlying behaviors that would contribute to
balling phenomenon\cite{lindstrom_simple_2023}. With this, CFD approximations or
surrogate modeling would suit as better options to modeling the potential
occurrence of balling, however are not integrated in this approach for previously
mentioned reasons.

\subsection{Model Context Protocol}
The Model Context Protocol (MCP) is a standard introduced by
Anthropic\cite{anthropic_anthropic_2025} which provides guidelines for
used functionality such as tool calling and resource querying;
Particularly suitable for agentic tasks\cite{anthropic_introducing_2024,
noauthor_modelcontextprotocolpython-sdk_2025, ehtesham_survey_2025,
hou_model_2025}. Along with these features, MCP is compatible with the wider
ecosystem of LLM providers with integrations in Software Development Kits (SDKs)
developed by Anthropic\cite{noauthor_modelcontextprotocolpython-sdk_2025},
OpenAI\cite{noauthor_model_nodate}, Google\cite{hamsa_buvaraghan_mcp_2025}, and
others\cite{hou_model_2025}.

\subsubsection{Tools}
Tools are the primary means in which an LLM is able to perform actions via MCP,
invoking external Application Programming Interfaces (APIs) or function calls
with the user's approval \cite{hou_model_2025,
noauthor_modelcontextprotocolpython-sdk_2025, ehtesham_survey_2025}. Complexity
of these tool calls can range from simple stateless actions (network requests to
external APIs) or long-lived function operations (simulation
runs)\cite{noauthor_modelcontextprotocolpython-sdk_2025}. Using the MCP Python
SDK\cite{noauthor_modelcontextprotocolpython-sdk_2025}, these tools can be
exposed to the LLM with the appropriate decorator function configuration as seen
in \ref{apx:example_tool}. A minimal configuration consisting of the decorator
wrapping a tool function is sufficient to register a tool for an LLM. However,
additional annotations such as type declarations, docstring descriptions, and
structured outputs significantly improve the model's comprehension and proper
utilization of the defined tool.

\subsubsection{Resources}
Resources provide the LLM capability to search for relevant information
regarding a given task. Similar to GET requests established within the Hypertext
Transfer Protocol (HTTP)\cite{noauthor_modelcontextprotocolpython-sdk_2025},
this allows for the query of information stored within databases, filesystems,
or other means to be accessible to the LLM\cite{hou_model_2025,
ehtesham_survey_2025}. This is particularly useful during the tool calling
process as it provides environmental state information regarding around the
model. However, during experiments the LLM's utilization of the resource
functionality proved unreliable and tools implementing the same functionality of
each resource displayed consistently reliable responses.

\begin{figure}[htbp]
    \centering
    \includegraphics[width=\textwidth]{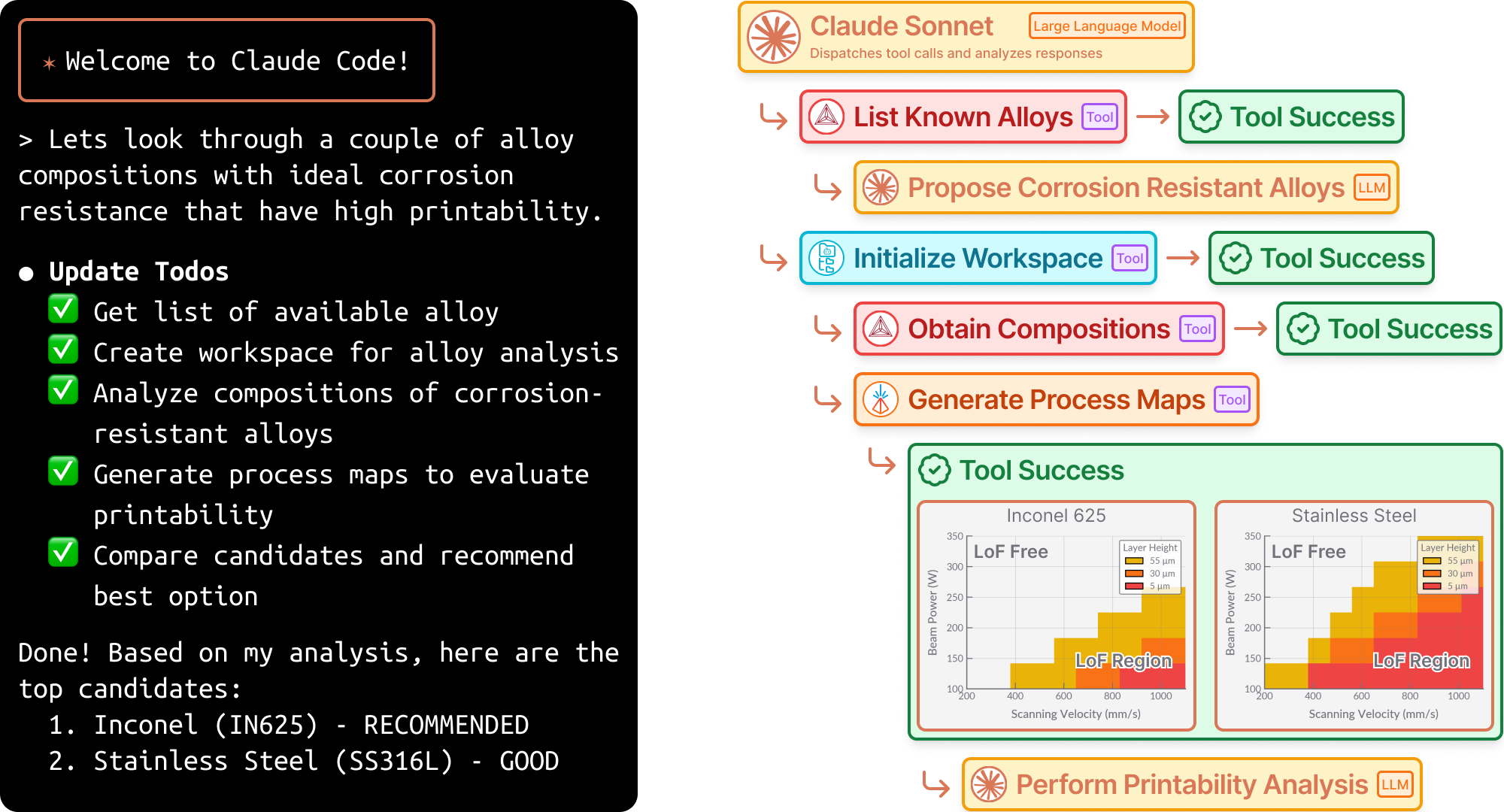}
    \caption{
    \textbf{(Left)} Claude Code provides an interface for integrating agentic
    tools with Claude Sonnet LLM, allowing for natural language input to execute
    tasks and response analysis. \textbf{(Right)} Streamlined summary of tool
    executions and analysis from the prompt given to Claude Code utilizing
    subagents for Additive Manufacturing, Thermo-Calc, and Workspace.
    }
    \label{fig:claude_code}
\end{figure}

\subsection{Agentic Tools}
Tools developed for this application are separated into three different MCP
Servers with Python Package Index (PyPi)\cite{noauthor_pypi_nodate} identifiers:
\texttt{additive-manufacturing}, \texttt{thermo-calc}, and
\texttt{workspace-agent}. Each of these packages maintain their own MCP server
tools and \textit{Subagent}\cite{noauthor_subagents_nodate} (a simple markdown
file providing additional context and system prompts to guide tool usage) and
are capable of standalone usage. Claude Code\cite{noauthor_claude_nodate}
provides a platform for coupling these tools into a multi-agent environment
utilizing Claude Sonnet 4\cite{noauthor_introducing_nodate} as the primary large
language model for orchestrating tool calls and performing response analysis. 

\subsubsection{Workspace}
The \texttt{workspace-agent} package, abbreviated to \texttt{wa} internally, is
responsible for the initialization and management of workspaces within the
context of tool calls. It primarily acts as a state management tool for storing
\texttt{JSON} serialized class objects and deserializng the stored \texttt{JSON}
files to use within tools. This approach allows for different tools to use the
same Python class methods and state since the input types to each tool are
limited to primitives accessible via command line. Thus, filename references to
serialized \texttt{JSON} files are provided as inputs to be loaded and
deserialized within each tool since \text{string}, \text{int}, and formatted
\text{list} and \text{dict} types are valid inputs to tools but Python instances
of classes or functions are not.

\begin{figure}[htbp]
    \centering
    \includegraphics[width=\textwidth]{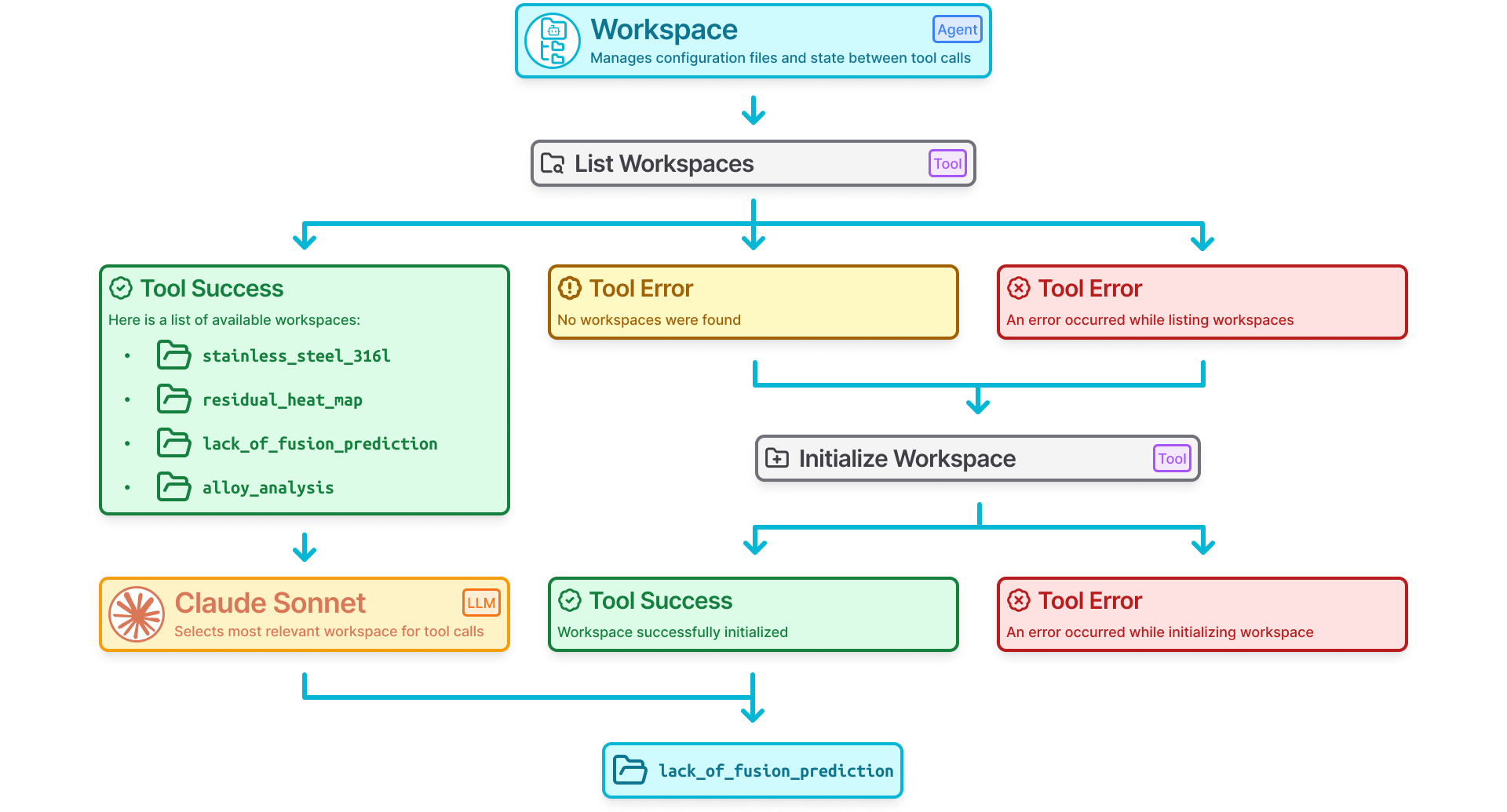}
    \caption{
    A simple tool calling procedure within the Workspace subagent for the task
    of finding or initializing a workspace. Here an initial tool call is made to
    list available workspaces and if none are found, a new workspace is created.
    This newly initialized workspace or the most relevant selected by Claude
    Sonnet is included in the successful response object.
    }
    \label{fig:workspace}
\end{figure}

Workspaces can be initialized via tool call (Figure \ref{fig:workspace}) and are
subdivided into top-level subfolders (i.e. \texttt{compositions},
\texttt{materials}, \texttt{process\_maps}, etc.) containing state and responses
for a given domain. Tools to list existing subfolders and their contents act as
guides to help the LLM navigate through potential tool input arguments narrow
the search window to the most applicable candidates. In addition to being listed
as tools, functionality to list subfolders and their content are also exposed as
resources accessible via MCP allowing for the LLM and user interface to obtain
this information implicitly via syntax (i.e.
\texttt{@workspace://\{workspace\}/\{subfolder\}/}).

The MCP server for workspaces operates independently, acting without direct
knowledge of other installed MCP servers relying only on the LLM context to
provide the appropriate arguments for workspace names and subfolders to
navigate. This architecture allows for the other MCP servers to utilize
workspace functionality preventing the issue of tool overlap between packages.
This opinionated approach allows for MCP servers that utilize
\texttt{workspace-agent} to share workspace subfolders such that the output of
one MCP server's tool can be utilized as an input to another MCP server tool.

\subsubsection{Thermo-Calc}
For this project, the authors developed and published a package with the
PyPI\cite{noauthor_pypi_nodate} identifier \texttt{thermo-calc} (aliased
internally as \texttt{tc}) to facilitate the installation and use of
Thermo-Calc's TC Python module\cite{thermo-calc_software_thermo-calc_2025}.
Along with these installation scripts, Command Line Interface (CLI) and MCP
bindings were also implemented for a subset of TC Python's functionality, those
of which are outlined in Section \ref{sec:thermo_calc_calcuation}. With the set
of tools managed by the thermo-calc subagent, the material properties of a given
alloy composition can be calculated and saved to a shared material configuration
that the tools from \texttt{additive-manufacturing} can then utilize for process
map generation.

\begin{figure}[htbp]
    \centering
    \includegraphics[width=\textwidth]{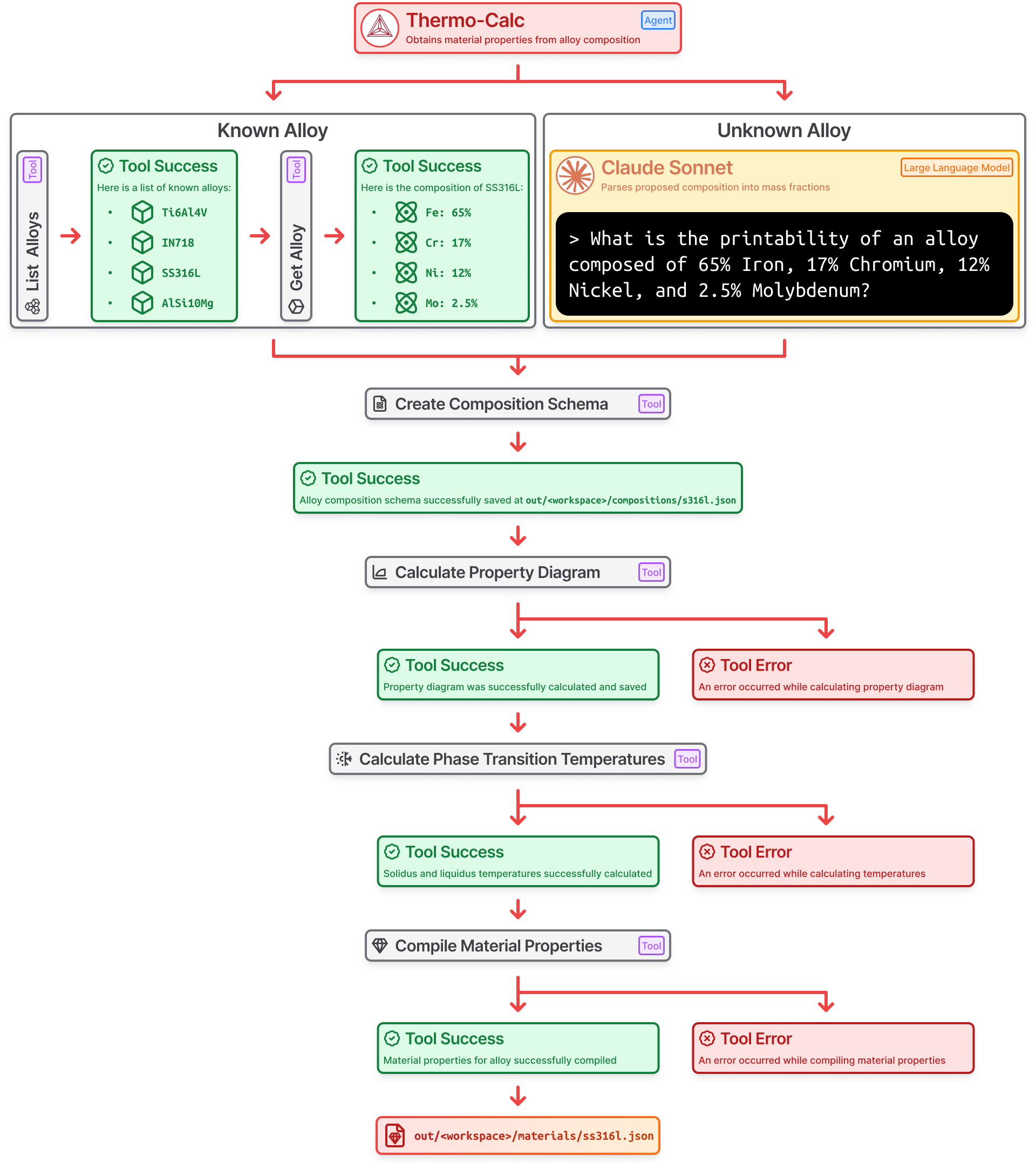}
    \caption{
    Flow diagram outlines the expected tool calling procedure for the
    Thermo-Calc subagent. In this example the material properties for Stainless
    Steel are extracted from the calculated property diagram of the alloy's
    elemental composition. Composition is obtained from a look-up table of known
    alloys or provided directly to the agentic system and parsed into mass
    fractions using Claude Sonnet. This process generates a schema file with
    alloy's material properties recorded for downstream use with other tools.
    }
    \label{fig:thermo-calc}
\end{figure}

In this process, an hypothetical alloy composition element or an existing element
name is provided to the large language model in order to create an alloy
composition file (Figure \ref{fig:thermo-calc}). An alloy composition file is
produced by serializing provided element proportions to a \texttt{JSON} file
consisting of element keys and mass fractions. (i.e. \texttt{\{"Fe": 0.9, "C":
0.1\}}) Both approaches eventually utilize the alloy composition schema tool,
however if the subagent is provided a generic alloy name, it utilizes tools to
list known alloys and obtain their compositions to then pass into the alloy
composition schema tool.

Once the alloy composition is obtained, the TC-Python API is called to
instantiate a server to calculate the Property Diagram for the given set of
elemental compositions and temperature range. This returns a
\texttt{PropertyDiagramResult} which is then saved for later use when
calculating phase transition temperatures and other material properties. In a
separate tool call, the \texttt{PropertyDiagramResult} is loaded for the volume
fraction calculation of the liquid phase which the liquidus and solidus phase
transition temperatures can be extracted from. The liquidus, solidus, and
melting (midpoint between liquidus and solidus) are then saved into a phase
transitions temperatures configuration file and utilized later in the
calculation of other material properties and serialization of the material
schema.

The previously calculated values are managed by \texttt{workspace-agent} package
and are saved in the \texttt{property\_diagrams},
\texttt{phase\_transition\_temperatures}, and \texttt{compositions} subfolders
respectively. The relevant configurations are loaded from these subfolders into
the material compilation tool which determines the values for the remaining
material properties, those being specific heat capacity, density, thermal
conductivity, and absorptivity. These values are instantiated as
\texttt{Material} class and serialized into the \texttt{materials} subfolder as
a \texttt{JSON} file for lack of fusion process map generation.

\subsubsection{Additive Manufacturing}
The \texttt{additive-manufacturing} (\texttt{am}) package provides a set of
tools to assist with the additive manufacturing build process, currently built
around providing feed forward solutions to predict potential build defects and
gauge its feasibility. Of the available features, the experimental setup
primarily utilizes the package's process map generation capabilities in order to
predict potential lack of fusion defect regimes. The exact process for computing
the melt pool dimensions necessary for determining the lack of fusion defect
regime is outlined in Section \ref{sec:defect_prediction}, but in short, melt
pool dimensions are calculated using an equation from Rosenthal
\cite{rosenthal_theory_2022} and fed into the lack of fusion defect criterion
(Equation \ref{eq:lof}) to classify the input process parameter combination. In
the current implementation classifications are limited to within or outside the
lack of fusion defect regime as the analytic solution only works for conduction
mode and cannot model melt pools in keyhole mode\cite{rosenthal_theory_2022}.

\begin{figure}[htbp]
    \centering
    \includegraphics[width=\textwidth]{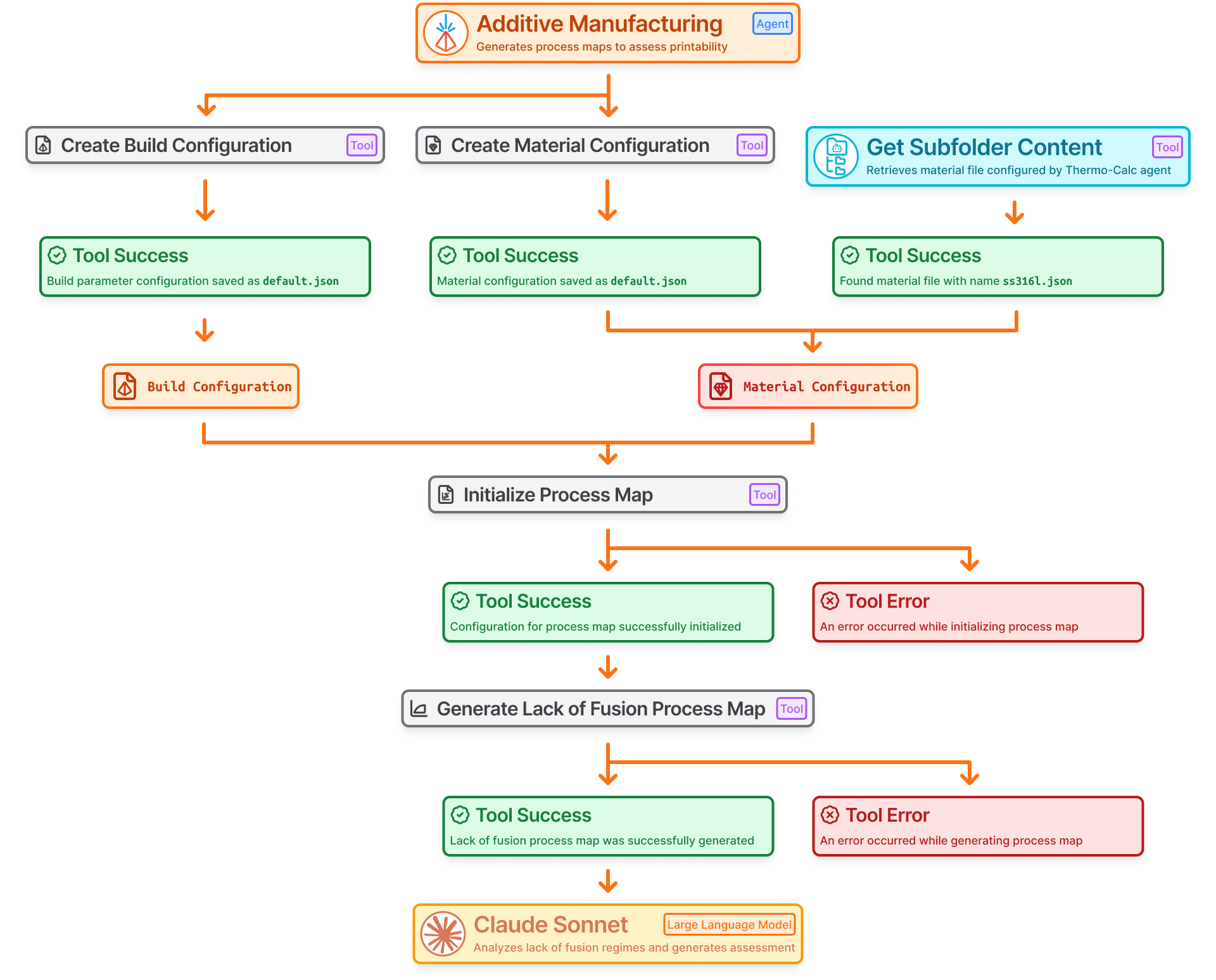}
    \caption{
    Diagram outlines expected tool calling procedure for additive manufacturing
    subagent for the task for generating and analyzing a lack of fusion process
    map. Build and material configurations are required to initialize a process
    map, the latter of which can be obtained from the Thermo-Calc subagent or
    manually configured by the additive manufacturing subagent. Initializing the
    process map provides override ranges for power and velocity build parameters
    for melt pool depth calculations. The tool generates the process map and a
    response consisting of power and velocity configurations that potentially
    exhibit lack of fusion defects. Claude Sonnet analyzes this response and
    provides suggestion for optimal build parameters.
    }
    \label{fig:additive_manufacturing}
\end{figure}

Process map generation is managed by the additive manufacturing subagent
responsible for the relevant tools within its MCP server (Figure
\ref{fig:additive_manufacturing}). To achieve this configuration files for
desired materials and build parameters need to be created before the process map
can be initialized and generated. Material configuration contains material
dependent properties such as density, thermal diffusivity, thermal conductivity,
and liquidus and solidus phase transition temperatures to name a few. Build
configuration manages values such as beam power, scan velocity, layer height,
and hatch spacing. These configuration files are utilized by the process map
initialization tool which creates the subfolder for storing process map results
and the process map configuration file for overriding build parameter
configurations of beam power and scan velocity. Instead of a single scalar
value, process map configuration file overrides the build configuration file
with a range of values as to use the same material and build configurations when
calculating defect regimes but with either a different power or velocity. The
default process map range for power and velocity is 100 to 1000 with steps every
100 W or mm/s respectively. With this, the necessary calculations are performed
to obtain lack of fusion regimes for 2 layer heights (-25 $\mu m$ and +25 $\mu
m$) in addition to the prescribed layer height within the provided range of
process parameters.

The Command Line Interface (CLI) provides the basic means of interaction with
the \texttt{additive-manufacturing} package, however with its MCP integration,
inputs and arguments can be left unstructured and functionality is accessible
through just natural language when utilized as tools within an MCP server. In
addition to natural language functionality, its integration with an LLM (i.e.
Claude Sonnet) augments the capabilities of the user enabling easy evaluation of
follow-up changes and LLM enabled analysis and feedback of computed results.
Through the \texttt{workspace-agent} package, material configurations generated
with the \texttt{thermo-calc} package can be utilized for the generation of
process maps, allowing for the end-to-end functionality of proposing alloy
compositions to analyzing their lack of fusion process map.

\subsection{Experiments}
Several experiments were performed to investigate the capability of the agentic
system. Each experiment consists of a simple prompt regarding the desired alloy
or material properties to explore (as shown in Figure \ref{fig:claude_code})
with the expectated generation of a lack of fusion process map along with
analysis and suggestion of potential printing parameters to implement. The whole
of these experiments aims to cover the wide range of potential use case
scenarios supported with the provided tools ranging from known and novel
alloy compositions to searching and assessing alloys with a desired material property
characteristic.

\subsubsection{Known Alloy Compositions}
Within the scope of known alloys, the system is expected to adhere to the
established approach to obtaining a process map for a given material
compositions with the caveat of obtaining these alloy composition from a
provided dictionary. In this task it is expected that the LLM only acts to
dispatch tools and interpret their subsequent response. This approach would be
evaluated on a subset of the known alloys which include: Stainless Steel 316L,
Titanium, Inconel, Aluminum, Tool Steel, Iron, Copper, Hastelloy X, K500,
Tungsten, Bronze, and Aluminum 7050.

\subsubsection{Material Property Search}
For an open ended task, such as that encountered during the search for an alloy
composition with specific desirable material properties, the large language
model is utilized more. In this case, a similar approach to that of the known
alloys is taken to determine an alloy's lack of fusion regime with an initial query given
to the LLM to provide a list of candidates with the desired material properties.
This included properties such as corrosion resistance, fatigue life, yield
stress, Young's modulus, fracture toughness, hardness, and ductility.

\subsubsection{Novel Alloy Compositions}
The LLM is able to interpret
a combination of elements into their respective mass fractions and call the
relevant tools to determine the alloy's lack of fusion regime. 
Compositions outside that of the known alloys were also included in this
investigation which primarily entailed the modification of existing alloy compositions such that an element was to be removed or the proportion of one was to be increased.
In addition, the agentic system also accepts an hypothetical combination of elements along with their prescribed proportions. This hypothetical combination of elements provides the freedom to combine various elements, such as 50\% Fe and 50\% N, into the system but in practice is limited by the bounds of thermodynamic calculations and databases of Thermo-Calc. The system was primarily evaluated with
compositions ranging from slight modifications of known alloys with a couple of completely
hypothetical combination of elements to element compositions proposed by the LLM.

\section{Results}

\begin{figure}[htbp]
    \centering
    \includegraphics[width=\textwidth]{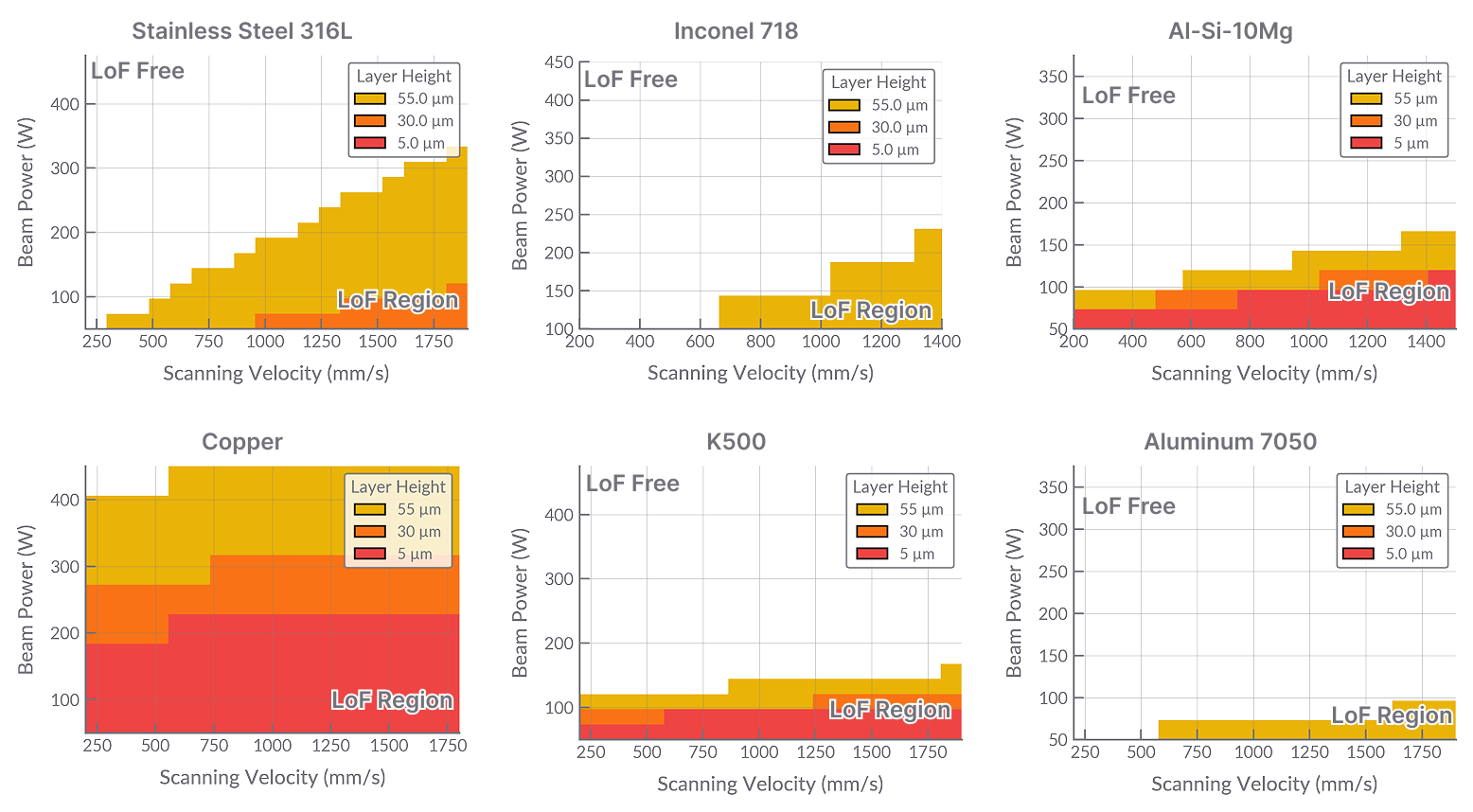}
    \caption{
    Lack of fusion process maps for a selection of known alloys with default
    hatch spacings of $50 \mathrm{\mu m}$ and various layer heights indicated by their respective colors within the legend.
    }
    \label{fig:alloys_known}
\end{figure}

For the known alloys, the lack of fusion process map was obtained using the
provided alloy composition. In these experiments, the large language model is
responsible for the selection of build parameters along with defining the
appropriate process map range within the power and velocity process parameters.
Of the 12 alloys this experiment was performed upon, 11 were able to produce a
suitable lack of fusion process map (Figure \ref{fig:alloys_known}). The one failure
in this case occurred with the Tool Steel prompt which resulted in the process
hanging while utilizing the property diagram calculation tool. This is potentially due to the number of potential options for "Tool Steel" that the MCP tool had available such as "D2 Tool Steel", "M2 Tool Steel", "A2 Tool Steel" which required further clarification.

\begin{figure}[htbp]
    \centering
    \includegraphics[width=\textwidth]{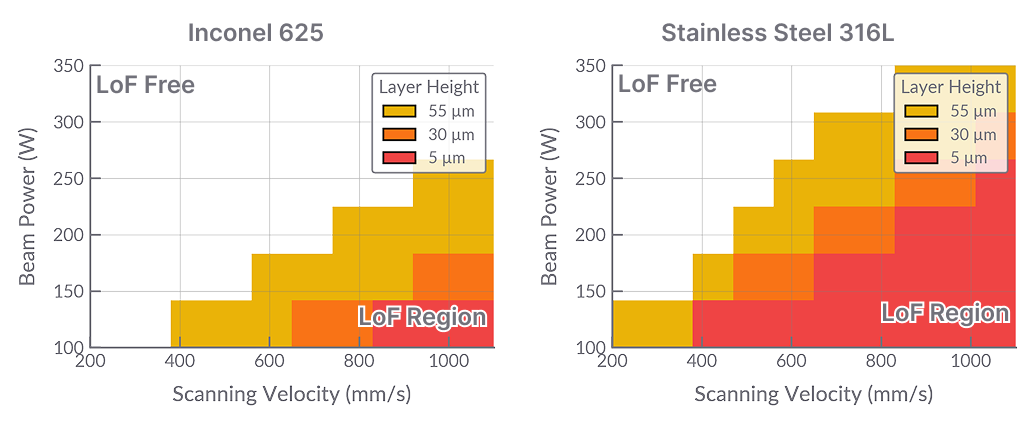}
    \caption{
        Comparison of lack of fusion process maps of between Inconel 625 and
        Stainless Steel 316L for corrosion resistant applications. Between these
        two candidates, the LLM ultimately recommended the use of Inconel 625
        for its smaller lack of fusion regime.
    }
    \label{fig:alloys_material_property_search}
\end{figure}

When prompted to search for an alloy with a specific material property, the LLM
reasons through number of potential candidates and selects a couple to evaluate.
The same procedure is taken for generating process maps for each alloy candidate
the results of which are evaluated and compared with emphasis towards
minimizing lack of fusion (Figure \ref{fig:alloys_material_property_search}). Of the 9
investigated material properties, the agentic system failed to produce process
maps for only one of the cases. The failure case occurred while comparing creep
resistance of various alloys, specifically when attempting to generate the
process map investigating the nickel based superalloy, Mar-M 247. Besides this
exception, all prompts were successful in creating lack of fusion process maps
and performing analysis and recommendations based on the feasibility of each alloy
candidate.

For the application of corrosion resistance (exact prompt and response provided in \ref{apx:llm_prompt_response}), the additive manufacturing subagent
recommendation of Inconel 625 aligns with the findings from multiple literature
sources\cite{zhang_achieving_2024, bolukbasi_improving_2023,
fan_mechanism_2024}. In a study by Zhang et al.\cite{zhang_achieving_2024}, the
authors investigate corrosion resistance of Inconel 625 and Stainless Steel 316L
hybrid alloys under the application of hydroflouric acid. Inconel is stated to
have greater corrosion resistant properties than that of Stainless Steel 316L
however, due to its higher Nickel and Chromium content the manufacturing cost is
higher as well\cite{zhang_achieving_2024}. Process parameter combinations used
with Inconel 625 range from 175 W and 500
mm/s\cite{gonzalez_characterization_2019} to 1000 W and 600
mm/s\cite{zhang_achieving_2024}, all of which are valid process parameters
within the prediction of the additive manufacturing subagent.

\begin{figure}[htbp]
    \centering
    \includegraphics[width=\textwidth]{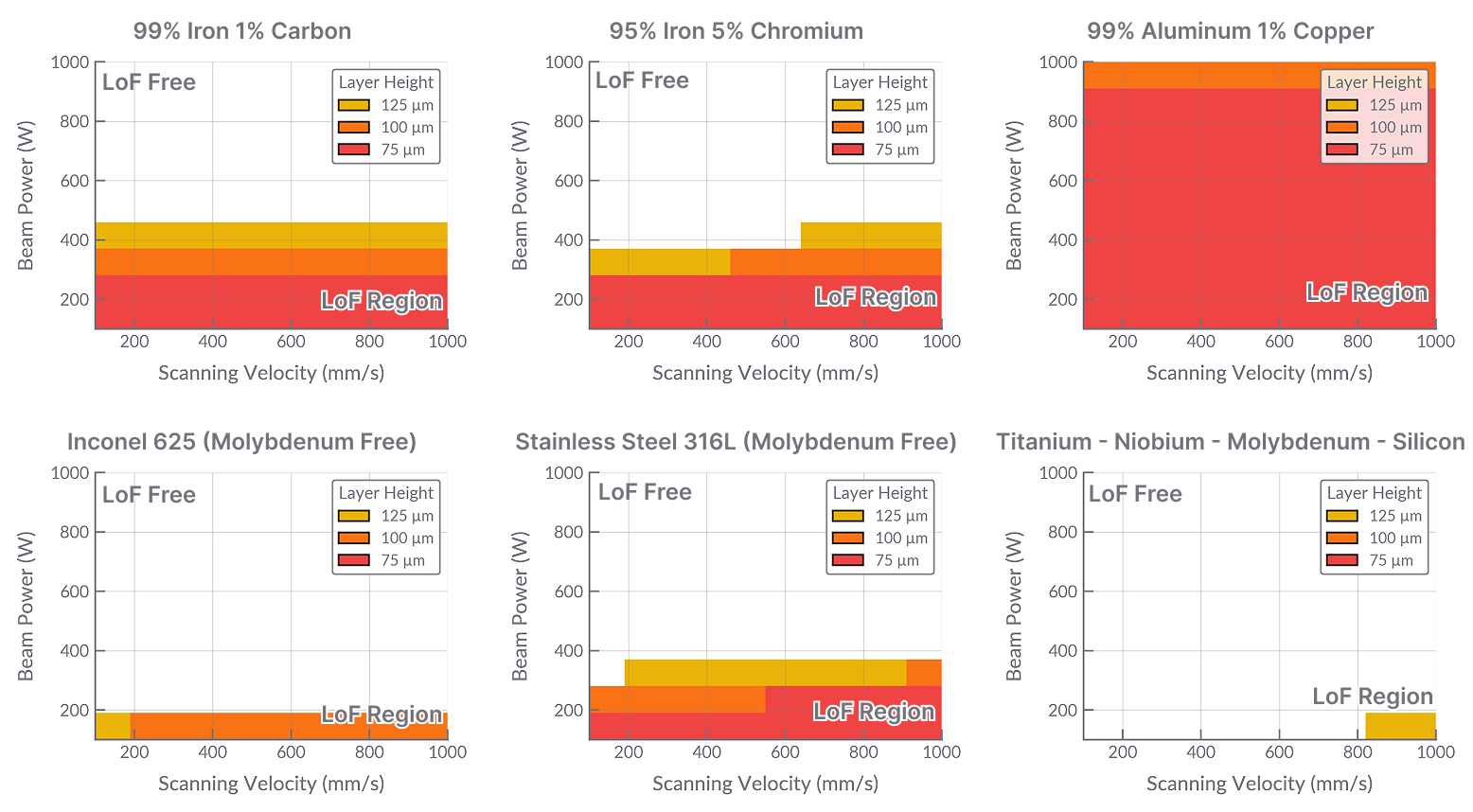}
    \caption{
    Various lack of fusion process maps for novel alloy compositions.
    \textbf{(Top Row)} Hypothetical element compositions for proposed Iron and
    Aluminum based alloys. \textbf{(Bottom Row)} slightly modified known alloy
    compositions of Inconel and Stainless Steel 316L with the removal of
    Molybdenum. \textbf{(Bottom Right)} An LLM proposed composition when prompted
    to suggest an novel alloy.
    }
    \label{fig:alloys_unknown}
\end{figure}

For novel alloys, compositions are directly evaluated and in some
instances produce rather extreme process maps. Of the 10 conducted experiments,
8 were successfully able to produce process maps with 2 failing due to the
nature of the alloy composition. Although the system accepts any composition of
elements, most hypothetical combination of elements are unsuitable for printing and
practical use. As such, slightly adjusting known alloy combinations produces
more suitable lack of fusion regimes as seen in the cases where Molybdenum is
removed for Stainless Steel 316L or Inconel 625 (Figure
\ref{fig:alloys_unknown}).

\subsection{Lack of Fusion Validation of Known Alloy Compositions}

Validation was performed on a couple of the common predicted alloy compositions
(IN718 and SS316L) to ensure the predicted lack of fusion process
aligned with what is found within the literature. In addition, the expected
material properties associated with each alloy along with the associated
processing parameters were included to provide broader context to what is
expected from each alloy.

\subsubsection{Inconel 718}
Validation of the lack of fusion predictions of known alloys indicate
satisfactory overlap between those generated by additive manufacturing subagent
and that found within published literature. In a study by Ghanadi et al.
\cite{ghanadi_effect_2025}, lack of fusion was observed at laser powers between
50 to 150 Watts and scan speeds of 700 mm/s to 1200 mm/s when conducted with a
prescribed layer height of $25 \mathrm{\mu m}$ and hatch space of $40
\mathrm{\mu m}$. These process parameters are similar to those implemented using
the additive manufacturing subagent ($50 \mathrm{\mu m}$ hatch spacing and $30
\mathrm{\mu m}$). Although the generated process map for Inconel 718 (Figure
\ref{fig:alloys_known}) denotes a slight under prediction of lack of fusion
compare to that of the literature, this can be attributed the more general
applicability of Rosenthal's equation\cite{rosenthal_theory_2022}.

The study provides areal surface roughness (Sa) of the top and sides of their printed lattice structure as an indicator for lack of fusion. The authors observed the expected trend that the Sa values decreased with increasing laser power and increasing scan speed, where the lowest Sa value of $26.5 \pm 5.4 \;\mu\text{m}$ with an ultimate shear strength of $153.2 \pm 7.7\;\text{Mpa}$ and shear strain of $0.33 \pm 0.02$ at 150 Watts and 200 mm/s\cite{ghanadi_effect_2025}. The highest Sa occurrence of $70 \pm 8.9 \mu\text{m}$ with an ultimate shear strength $8.93 \pm 0.9\;\text{Mpa}$ and $0.69 \pm 0.03$ shear strain recorded at a 50 Watts and 1200 mm/s power and velocity combination\cite{ghanadi_effect_2025}.

\subsubsection{Stainless Steel 316L}
The generated
process map for Stainless Steel 316L (SS316L) matches the lack of fusion regimes
outlined in the literature, with unfavorable process windows at lower powers and
higher velocity combinations\cite{ahmed_process_2022, huang_microstructure_2025,
parikh_property-graded_2023}. Ahmed et al.\cite{ahmed_process_2022} encounters
high densification of SS316L at power and velocity combinations of (150 W, 500
mm/s), (200 W, 700 mm/s), (250 W, 900 mm/s), and (300 W, 1100 mm/s), matching
that presented in Figure \ref{fig:alloys_known}.

Within the lack of fusion defect regime, the resulting pore size is heavily
dependent on the factors of height and width of the melt pool along with the
hatch spacing and layer height of the build\cite{cacace_lack_2022,
pak_thermopore_2024}. For hatch spacings ranging from 25 $\mu m$ to 55 $\mu m$,
equivalent pore sizes upwards of 160 $\mu m$ can be observed (power at around
100 W, 30 $\mu m$ layer height, and 1.4 m/s scan
velocity)\cite{pak_thermopore_2024}. Within the range of scan velocities from
around 1.05 m/s to 1.75 m/s, an average pore diameter of around 30 $\mu m$ with
a standard deviation of 16 $\mu m$ and approximate maximum of 130 $\mu m$ was
observed (with previously mentioned process parameters and 50 $\mu m$ hatch
spacing)\cite{pak_thermopore_2024}.

In more nominal processing conditions, the percentage of porosity within a part
is observed to change as a function of energy density. This is reflected in work
by Tucho et al. where the authors manufactured a number of samples and recorded
that porosity by volume within a part decreases from over 3\% with energy
densities of around $50\;\text{J/mm}^3$ to less than 1\% at energy densities of
$65\;\text{J/mm}^3$\cite{tucho_investigation_2018}. The authors also found that
hardness increases linearly with energy density as a hardness values of
$188\pm4$ HV was observed at an energy density of $80\;\text{J/mm}^3$ from a
value of $168\pm15$ HV at an energy density of
$50\;\text{J/mm}^3$\cite{tucho_investigation_2018}. With regard to mechanical
properties, it was found that tensile strength and yield strength could be
improved through layer remelting as Lu et al. recorded a respective increase
from 674 MPa to 725 MPa and 591 MPa to 643 MPa with process parameters of power
at 250 W, scan velocity at 950 mm/s, hatch spacing of 110 $\mu m$, and layer
height of 30 $\mu m$\cite{lu_research_2020, ahmed_process_2022}.



\section{Discussion}

In this agentic system, all tasks are completed with tools performing
deterministic actions; the large language model merely orchestrates and analyses
the response of dispatched tool calls. As a result, fidelity of the system's
predictions relies primarily on the accuracy of the developed tools. To that
point, this work limits its process map prediction to the lack of fusion regime
as it utilizes a conduction mode limited analytical approach to obtain melt pool
dimensions\cite{rosenthal_theory_2022}. With these separation of concerns,
determining the source of invalid predictions is simplified where incorrect
evaluation of results can be attributed to the large language model and
production of inaccurate predictions can be assigned to the tool.

\begin{figure}[htbp]
    \centering
    \includegraphics[width=\textwidth]{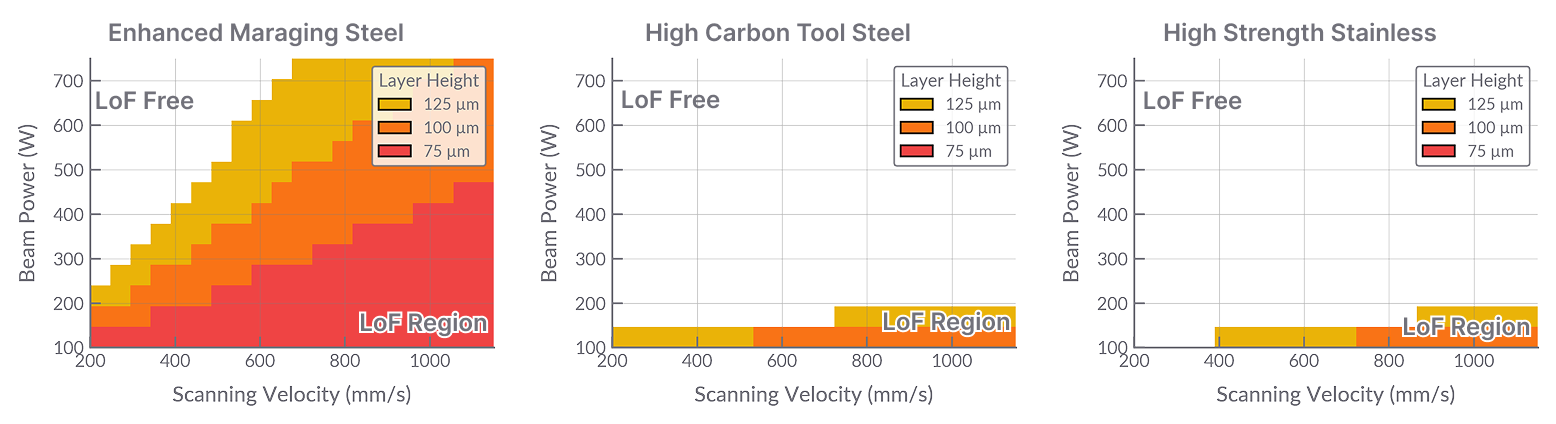}
    \caption{
    High hardness alloy candidates with their respective lack of fusion process
    maps. LLM makes an error by recommending Enhanced Maraging Steel over High
    Strength Stainless Steel for its smaller lack of fusion region.
    }
    \label{fig:alloys_wrong}
\end{figure}

Within the experimental trials a few prompts produced incorrect results as such
was the case in evaluating a material with optimal hardness. In this case the
lack of fusion process maps were correctly generated however the LLM recommended
the use of Enhanced Maraging Steel over the other two candidates (Figure
\ref{fig:alloys_wrong}). From visually interpreting the process maps, it is
clear that Enhanced Maraging Steel has the largest lack of fusion regime when
compared to its counterparts. One potential cause of this misinterpretation
could be attributed to the response data structure returned to the LLM after a
tool call. Since the lack of fusion regimes are returned as power and velocity
combinations within a list of tuples (i.e. \texttt{[(100, 100), (150, 100),
...]}), the LLM may have misinterpreted these values to be valid process windows
rather than lack of fusion regimes. Proper key names and serializing of results
would help reduce these types of LLM mistakes.

The deterministic nature of these tools does present the question of the benefit
of introducing an LLM into the system, as if the tools can work on their own
individually, what would be the purpose of including an LLM into the
architecture. This is a valid concern as the common workflows of each tool can
be connected to each other through a parent script, achieving the same effect.
However, the inclusion of an LLM and the separation of tools into their own
subagents allows for several unique capabilities. The most apparent of these is
the utilization of natural language when interacting with the developed tools.
This allows the user to directly interact with the available tooling without
explicitly adhering to the strict syntax and argument guidelines required by the
CLI. Another is the LLM's ability to analyze results providing user interpretive
feedback on a tool's response. This supports another key ability of LLM agents
which is to establish a feedback loop and respond to changes within its context.
The integration of LLM Agents within the developed tools not only enhances the
user experience with these tools but establishes an automation framework enabling
researcher to efficiently utilize their available tools.

\subsection{Comparison to Existing Tool Integration Frameworks}
Other alternatives to tool calling and agentic system frameworks exist to the Model Context Protocol (MCP) such as LangChain\cite{noauthor_langchain_nodate} and LlamaIndex\cite{noauthor_llamaindex_nodate}, however, these approaches lack first party integration into the LLM that MCP provides. The alternative approaches are no less capable than MCP and have their own specific features as seen in other various agentic works \cite{jadhav_llm-3d_2025, chandrasekhar_amgpt_2024}. The primary reasoning behind using MCP as the framework for designing the agentic tools is expected long term support and ease of use that the protocol provides when applied to other models and the broader support it has with providers such as Anthropic, Google, and OpenAI. Although much of these tests were performed with the Claude model from Anthropic, initial implementations show that these experiments can be performed on the latest models from OpenAI (ChatGPT) and Google (Gemini) as well.

\section{Conclusion}

This multi-agent system augments the researcher's ability to investigate and
evaluate known and novel alloy compositions for processing via additive
manufacturing. This is achieved through invoking tool calls via the Model
Context Protocol (MCP) to perform thermophysical calculations and melt pool
dimensional approximations to generate a lack of fusion process map. In
addition, these tools can leverage the generalized knowledge of LLMs to assist in more generalized tasks such as suggesting alloy compositions for desired material properties through
querying its large base of knowledge. The predictions from this system provide a suitable starting point of process parameters for experimental trials that would avoid potential areas of lack of fusion. Expansions upon this work would explore a more comprehensive search for processing parameters that would also consider the keyholing and balling defect regimes. This work sets
the foundation for complex tool usage and automated research within the additive
manufacturing field.

\section{Future Work}

Although the authors have provided a comprehensive study on the use of agentic
tools within the additive manufacturing space, there remain a couple of aspects
to this work that would benefit from further research.

\subsection{Technical Roadmap}
\subsubsection{Objective}
Extension upon this work would expand upon the generation of a robust process
map which considers additional build parameter and processing factors such as
beam size, melt pool fluid dynamics, and varying absorptivity. Specifically,
process regimes for keyholing and balling would be incorporated into the final
alloy evaluation when selecting a set of build parameters within a process
window. Additionally, validation of novel alloy compositions and process
parameters proposed by the agentic system would include fabricating samples
evaluation through tensile tests.

\subsubsection{Methodology}
In order to obtain melt pool dimensions which more accurately match those
created experimentally, a more comprehensive solver will need to be utilized.
Options include CFD solvers developed by FLOW-3D \cite{noauthor_flow-3d_nodate},
OpenFOAM\cite{noauthor_openfoam_2025}, or one specifically for the purpose of
defining defect regimes. With each of these approaches, the tradeoff between
computational cost and accuracy would be of primary concern.

Within the agentic system, the overall framework will still build upon the Model
Context Protocol and further utilize features such as resources and tool
calling. Design improvements would include tool consolidate to reduce potential
tool bloat and clarify adjustments to reduce the frequency of faulty tool calls.
Additionally, consideration towards optimizing token and time cost between
various LLM providers will be incorporated as well.

\subsubsection{Evaluation Metrics}
Evaluation of the updated solver will compare the melt pool dimensional values
to those observed within the literature. Metrics here will include the percent
error observed between the two dimensional measurements. Additionally, the
solver's ability to model behaviors such as balling and keyholing will be
scrutinized further and the computed defect regimes will be contrasted to those
found within the literature.

Sample fabrication would have a more qualitative evaluation metric as machine or
composition specific factors may hinder the manufacturing of parts consisting of
novel alloy compositions. A scenario where the novel composition sample is
ultimately fabricated but under a certain process conditions not mentioned by
the agentic system. After fabrication, the mechanical properties such as yield
strength, ultimate tensile strength, and hardness would be evaluated as well.

\subsubsection{Timeline}
The objectives outlined within this technical roadmap such as improvements to
the agentic system are currently in progress and will be expanded upon further
in a future work. Other considerations such as part fabrication and mechanical
testing relies on the selection or development of a CFD solver that can produce
more accurate melt pool dimensions suitable for a more comprehensive process
map.

\section{Data Availability}
The multi-agent system developed for this work along with the data associated
with experimental prompts are available at the following link:
\url{https://github.com/BaratiLab/Agentic-Additive-Manufacturing-Alloy-Evaluation}.
The various tools within the agentic system are hosted on the Python Package
Index (PyPI) and installable with their respective package names:
\texttt{additive-manufacturing}, \texttt{thermo-calc}, \texttt{workspace-agent}.
\clearpage
\appendix

\renewcommand{\thesection}{Appendix \Alph{section}}

\section{Model Context Protocol Example Tool}
\label{apx:example_tool}

\begin{figure}[h!]
\centering
\begin{minipage}{\linewidth}
\begin{lstlisting}[style=mypython]
@app.tool(
    title="Run Layer with Solver",
    description="Runs solver on a segments file (segments file should be one layer) and saves the generated meshes.",
    structured_output=True,
)
def solver_run_layer(
    workspace: str,
    segments_foldername: str,
    layer_number: int,
    build_config_filename: str = "default.json",
    material_config_filename: str = "default.json",
    mesh_config_filename: str = "default.json",
    run_name: str | None = None,
) -> Union[ToolSuccess[Path], ToolError]:
    """
    Runs solver for segments at a specified layer number.
    Args:
        workspace: Folder name of existing workspace
        segments_foldername: Folder name of where segments are expected to be found.
        layer_number: Layer number to run solver on, typically starts from 1. For testing out, try skipping the first several layers as those sometimes don't include part geometry.
        distance_xy_max: Maximum segment length when parsing (defaults to 1.0 mm).
        build_config_filename: build config file to use with solver.
        material_config_filename: material config file to use with solver.
        mesh_config_filename: mesh config file to use with solver.
        run_name: Name of folder to save generated meshes at, typically autogenerated.
    """
    ...
\end{lstlisting}
\caption{Example tool implementation shown via Solver tool defined in additive-manufacturing package.}
\label{fig:mcp_tool_example}
\end{minipage}
\end{figure}

\section{Melt Pool Length}
\label{apx:melt_pool_length}
Rosenthal's equation\cite{rosenthal_theory_2022} can be rewritten to provide
the bounds for the melt pool. If $z$ is substituted with $R$, the width of the
melt pool shrinks to zero and becomes the furthest point in the melt pool.

\begin{equation}
z = R + \frac{2 \alpha}{V} ln\left(\frac{2 \pi k R \Delta T}{\epsilon P}\right)
\label{eq:rosenthal_length_1}
\end{equation}

\begin{equation}
R = R + \frac{2 \alpha}{V} ln\left(\frac{2 \pi k R \Delta T}{\epsilon P}\right)
\label{eq:rosenthal_length_2}
\end{equation}

\begin{equation}
0 = \frac{2 \alpha}{V} ln\left(\frac{2 \pi k R \Delta T}{\epsilon P}\right)
\label{eq:rosenthal_length_3}
\end{equation}

\noindent Since the coefficient term $\frac{2\alpha}{V}$ will be a non-zero
value, ignore this term to simplify calculation.

\begin{equation}
\frac{2\alpha}{V} \neq 0;\;0 = ln\left(\frac{2 \pi k R \Delta T}{\epsilon P}\right)
\label{eq:rosenthal_length_4}
\end{equation}

\noindent Taking the exponential of both sides and solving for $R$ produces the
following equation that provides the expected length of the melt pool from the
heat source.

\begin{equation}
e^{ln\left(\frac{2 \pi k R \Delta T}{\epsilon P}\right)} = e^0
\label{eq:rosenthal_length_5}
\end{equation}

\begin{equation}
\frac{2 \pi k R \Delta T}{\epsilon P} = 1
\label{eq:rosenthal_length_6}
\end{equation}

\begin{equation}
2 \pi k R \Delta T = \epsilon P
\label{eq:rosenthal_length_7}
\end{equation}

\begin{equation}
R = \frac{\epsilon P}{2 \pi k \Delta T}
\label{eq:rosenthal_length_8}
\end{equation}
\section{Database Selection}
\label{apx:database_selection}
\begin{enumerate}
    \item \textbf{Ti-based alloys:} if \texttt{top element is "Ti"} $\rightarrow$ \texttt{TCTI6}.
    \item \textbf{Ni-based alloys:} if \texttt{top element is "Ni"} or the Ni fraction is $\ge 0.30$ $\rightarrow$ \texttt{TCNI12}.
    \item \textbf{Fe-based alloys:} if \texttt{top element is "Fe"} (and the previous Ni criteria were not satisfied) $\rightarrow$ \texttt{TCFE14}.
    \item \textbf{Al-based alloys:} if \texttt{top element is "Al"} (and none of the above criteria applied) $\rightarrow$ \texttt{TCAL9}.
    
     \item \textbf{Pure elements:} if we have a pure metal which is just a single element$\rightarrow$ \texttt{PURE5}.

     \item \textbf{Fallback:} if none of the conditions above are met (maybe for MPEAs) $\rightarrow$ \texttt{TCHEA7}.
\end{enumerate}
\begin{figure}[h!]
    \centering
    \includegraphics[width=0.55\textwidth]{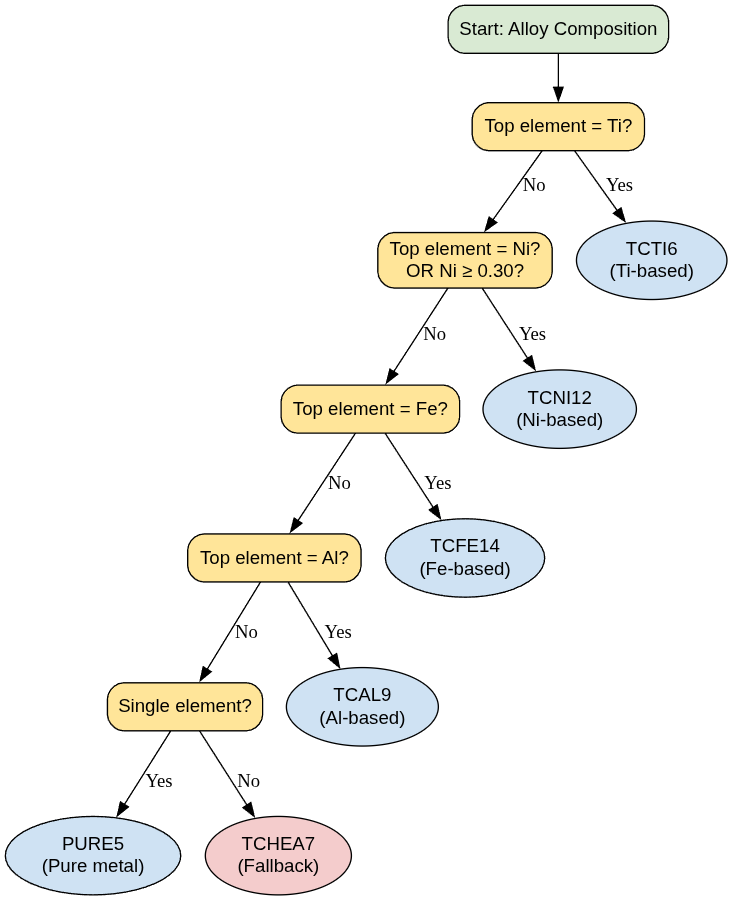}
    \caption{Database selection decision tree}
\end{figure}
\newpage

\section{Validation of Predicted Absorptivity}
\label{apx:validation_of_absorptivity}
\begin{table}[h!]
\centering
\begin{tabular}{lccc}
\hline
\textbf{Alloy} & \textbf{Reported Absorptivity} & \textbf{Drude-Model Absorptivity} \\
\hline
SS316L & 0.35-0.60 & 0.4479 \\
Ti-6Al-4V & 0.27-0.6 & 0.4520 \\
Hastelloy X & 0.45--0.64 & 0.4413 \\
M300 Steel & 0.35 & 0.43 \\
\hline
\end{tabular}
\caption{Comparison between reported (literature) absorptivity and Drude-model calculated absorptivity at 1070~nm.}
\end{table}

\section{Lack of Fusion and Keyhole Boundaries of Known Alloys}
\label{apx:lack_of_fusion_keyhole_mode_boundaries}

\begin{figure}[htbp]
    \centering
    \includegraphics[width=\textwidth]{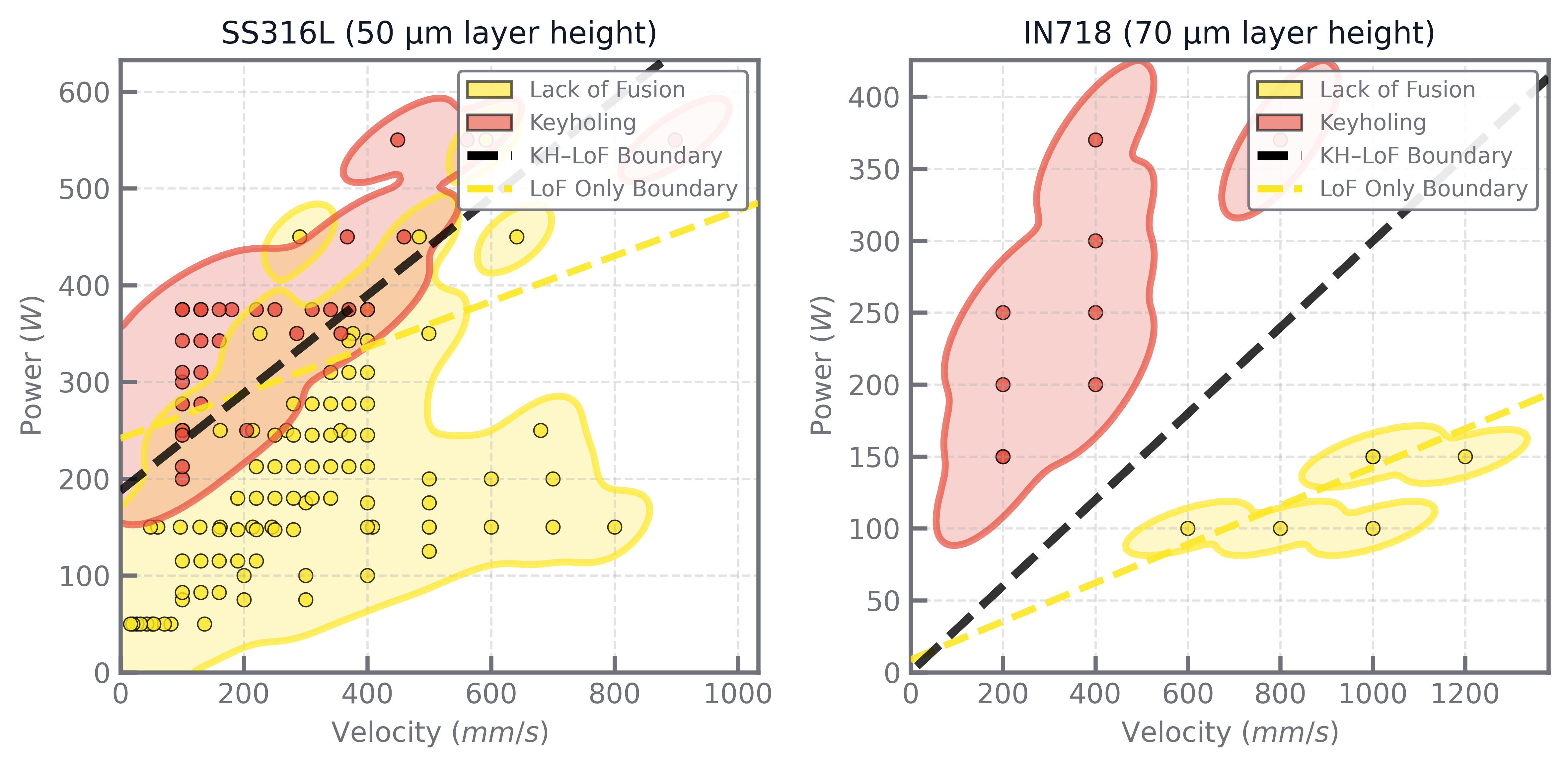}
    \caption{
    Literature values for SS316L and IN718 melt pool classifications expected
    defect regions for each respective alloy. Logistic regression is applied to
    establish a boundary between the two regions to reference as threshold to
    the extent in which Rosenthal's equation can be used to predict lack of
    fusion defects.
    }
    \label{fig:keyhole_lof_boundary}
\end{figure}

Literature results for defect classifications regarding 316L Stainless Steel and
Inconel 718 were analyzed to uncover potential lack of fusion and keyhole
regions over a range of power and velocity
combinations.\cite{akbari_meltpoolnet_2022, tapia_gaussian_2018}. Logistic
Regression was applied to determine boundary between the keyholing and lack of
fusion regions. With respect to SS316L, a higher degree of overlap between these
two defect regimes is visible due to the greater amount of samples when compared
to that of the IN718 plot (Figure \ref{fig:keyhole_lof_boundary}).

The boundary set by logistic regression implies that given the two defection
classifications, if a defect were to occur it would be either keyhole or lack of
fusion since it is expected that between these two regions a nominal process
window could exist. This boundary is utilized as a threshold to limit the extent
in which the Rosenthal's equation can be applied to determine lack of fusion
defects before exiting conduction mode. In addition, sensitivity analysis on the
boundary was performed by removing data points with the keyhole classification
and the boundary was calculated only with lack of fusion data points. This
resulted in a more conservative boundary in which defects would be classified as
lack of fusion.

\section{LLM Prompt and Response for Corrosion Resistant Material Property Search}
\label{apx:llm_prompt_response}

The following prompt was provided to the agentic system:

\begin{quote}
Lets look through a couple of alloy compositions to find a suitable candidate ideal for corrosion resistance that express high printability?
\end{quote}

\noindent{After successful tool calls, the LLM responded with the following:}

\begin{figure}[h!]
\centering
\begin{minipage}{\linewidth}
\begin{lstlisting}[style=mypython]
Based on my analysis of corrosion-resistant alloys, here are the top candidates:

  IN625 (Inconel 625) - RECOMMENDED

  Composition: 62.5% Ni, 22.9% Cr, 9.4% Mo, 5.2% Fe
  - Corrosion Resistance: Excellent in marine, chemical, and high-temperature environments
  - Printability: Superior - Only 3 lack-of-fusion conditions at low power (100W) with high velocities
  - Process Window: Wide range of successful parameters (150-400W, 200-1100 mm/s)

  SS316L (Stainless Steel 316L) - GOOD ALTERNATIVE

  Composition: 67.4% Fe, 17.6% Cr, 12.4% Ni, 2.6% Mo
  - Corrosion Resistance: Very good, especially against chloride environments
  - Printability: More challenging - 17 lack-of-fusion conditions at low power, 38 at higher layer heights
  - Process Window: Narrower than IN625, requires careful parameter control

  Key Findings:

  1. IN625 shows superior printability with fewer defect-prone parameter combinations
  2. SS316L has more extensive lack-of-fusion regions, particularly at 55μm layer height
  3. Both alloys offer excellent corrosion resistance, but IN625 provides better process reliability

  Recommendation: IN625 is the optimal choice, combining exceptional corrosion resistance with superior additive manufacturing printability.
\end{lstlisting}
\caption{Prompt response and reasoning provided by agentic system for evaluating lack of fusion process map for known alloys.}
\label{fig:llm_prompt_response}
\end{minipage}
\end{figure}

\bibliography{references}

\end{document}